\documentclass{article}

\usepackage{microtype}
\usepackage{graphicx}
\usepackage{subfigure}
\usepackage{booktabs} % for professional tables

\usepackage{hyperref}

\usepackage[accepted]{icml2023-diffxyz}

\usepackage{amsmath}
\usepackage{amssymb}
\usepackage{mathtools}
\usepackage{amsthm}
\usepackage{bm}
\usepackage{bbm}
\usepackage{mathtools}
\usepackage{amsthm, soul}
\usepackage[capitalize,noabbrev]{cleveref} % if you use cleveref
\usepackage{listings}
\usepackage{xcolor}
\usepackage[font=small,skip=5pt]{caption}
\usepackage{enumitem}

\definecolor{codegreen}{rgb}{0,0.6,0}
\definecolor{codegray}{rgb}{0.5,0.5,0.5}
\definecolor{codepurple}{rgb}{0.58,0,0.82}
\definecolor{backcolour}{rgb}{0.95,0.95,0.92}

\lstdefinestyle{mystyle}{
    backgroundcolor=\color{backcolour},   
    commentstyle=\color{codegreen},
    keywordstyle=\color{magenta},
    numberstyle=\tiny\color{codegray},
    stringstyle=\color{codepurple},
    basicstyle=\ttfamily\footnotesize,
    breakatwhitespace=false,         
    breaklines=true,                 
    captionpos=b,                    
    keepspaces=true,                 
    numbers=left,                    
    numbersep=5pt,                  
    showspaces=false,                
    showstringspaces=false,
    showtabs=false,                  
    tabsize=2
}
\theoremstyle{plain}

\theoremstyle{definition}

\theoremstyle{remark}

\usepackage[textsize=tiny]{todonotes}
\usepackage[capitalize,noabbrev]{cleveref}

\def\argmin{\mathop{\mathrm{arg\,min}}}
\def\xbm{{\bm{x}}}
\def\ybm{{\bm{y}}}
\def\Abm{{\bm{A}}}
\def\R{\mathbb{R}}

\def\eg{\emph{e.g.}}

\icmltitlerunning{Differentiable CT Projector}

\begin{document}

\twocolumn[
\icmltitle{Differentiable Forward Projector for X-ray Computed Tomography}



\icmlsetsymbol{equal}{*}

\begin{icmlauthorlist}
\icmlauthor{Hyojin Kim}{equal,llnl}
\icmlauthor{Kyle Champley}{equal,ziteo}
\end{icmlauthorlist}

\icmlaffiliation{llnl}{Center for Applied Scientific Computing, Lawrence Livermore National Laboratory, California, USA}
\icmlaffiliation{ziteo}{Ziteo Medical, California, USA, He conducted the software development and experiments while affiliated with Lawrence Livermore National Laboratory}

\icmlcorrespondingauthor{Hyojin Kim}{hkim@llnl.gov}

\icmlkeywords{Machine Learning, Computed Tomography, Differentiable Projector, ICML}

\vskip 0.3in
]



\printAffiliationsAndNotice{\icmlEqualContribution} % otherwise use the standard text.

\begin{abstract}
Data-driven deep learning has been successfully applied to various computed tomographic reconstruction problems. The deep inference models may outperform existing analytical and iterative algorithms, especially in ill-posed CT reconstruction. However, those methods often predict images that do not agree with the measured projection data. This paper presents an accurate differentiable forward and back projection software library to ensure the consistency between the predicted images and the original measurements. The software library efficiently supports various projection geometry types while minimizing the GPU memory footprint requirement, which facilitates seamless integration with existing deep learning training and inference pipelines. The proposed software is available as open source: \url{https://github.com/LLNL/LEAP}. 
\end{abstract}

% introduction
\section{Introduction}

X-ray Computed Tomography (XCT) is a 3D noninvasive/nondestructive imaging modality that has applications in healthcare, security, and industry \cite{Martz2016}.  X-ray projections are recorded by an x-ray detector opposite of an x-ray source over a range of angles.  The XCT inverse problem attempts to reconstruct the 3D volume of x-ray linear attenuation coefficients from this collection of x-ray projections. The XCT reconstruction problem has been extensively studied for several decades. Recently, deep learning (DL)-based inference models have successfully tackled more challenging reconstruction problems, such as scenarios where x-ray projections are only collected over a small range of angles (limited-angle CT) or where few projections are collected (sparse-view or few-view CT) which are characterized as ill-posed inverse problems \cite{han2018unet,zhou2020diffraction,zhang2020b,zhang2021d,gao2022,hu20224d}. 

While recent advancements in deep neural network trained models have been shown to outperform existing numerical optimization methods in some applications, they often suffer from a lack of data consistency to check whether the predicted images agree with the measured X-ray projection data (sinogram). Moreover, those inference models rely mainly on the training data distribution with the loss functions typically based on the reconstruction error between the predicted images and the ground-truth images. In these XCT inverse problems, it is crucial to incorporate the data consistency step into the inference models to suppress potential artifacts and hallucinated regions. To enable this, a proper forward model needs to be incorporated into the training pipeline or the inference step. 

One way to facilitate this is the use of an additional data consistency step with a loss function in the training or inference pipeline. The loss functions typically aim to minimize the difference between the forward-projected data of the predicted volume and the original projection data \cite{zhou2019LACT, liu2022DOLCE, lahiri2023}. In this setup, a differentiable forward projection model needs to be integrated into the neural network framework. Another way is to use the predicted volumes from the inference models as a prior (initial guess) in additional reconstruction algorithms \cite{kim2019FVCT, anirudh2019improving, zhou2020diffraction}. Similarly, the inferred volumes are forward projected to complete the projection data in limited-angle CT applications \cite{anirudh2018LACT}. Although the additional reconstruction algorithms and the sinogram completion step can be performed separately, an integrated end-to-end pipeline including the neural network inference models and following reconstruction processes make the application more cohesive and practical. 

Despite the increased demand for differentiable projectors in a wide range of XCT applications, existing solutions for differentiable projections are somewhat limited and impractical. The Radon Transform is widely used, but works only with the parallel beam geometry. One could pre-compute and store the projection matrix for any desired CT geometry \cite{lahiri2023}, but this method utilizes an enormous amount of memory (even though it is a sparse matrix) and is significantly inefficient because fetching the system matrix values from memory is much slower than computing these coefficient on the fly. In the case of cone-beam CT or high-resolution image reconstruction, the matrix size becomes excessively large, which makes the forward projection impractical. A neural network-based projector method \cite{gupta2018cnnproj} is also less practical, due to the training requirement for each geometry. Most CT systems have a configurable geometry to optimize image quality and it is impractical to retrain with each configuration. The absence of a widely applicable, memory-efficient differentiable CT projection software tool remains a significant limitation.

This paper presents a new software library tool called LivermorE AI Projector (LEAP), providing differentiable CT forward projections to be easily integrated into neural network-based inference models. This tool provides both CPU- and GPU-based forward and back projections, utilizing a significantly small memory footprint, particularly with extensive GPU memory demands during neural network training and inference time. The main contributions are summarized as:

\begin{itemize}[noitemsep,topsep=0pt]
\item The proposed tool provides differentiable projections to enable an end-to-end neural network training or inference pipeline for CT applications, without the need for significant memory footprint requirements. 
\item This supports forward and back projections for three widely used 3D scanner geometry types: parallel-beam, cone-beam, and a method to specify arbitrary locations and orientations of a set of source/detector pairs.
\item This utilizes the highly accurate Separable Footprint (SF) projector model. Our implementation is also quantitatively accurate and all numerical values scale appropriately when changing the voxel sizes, detector sizes, etc.
\item The proposed differentiable CT projections leverage the PyTorch interface and tensor formats, enabling seamless integration into existing neural network frameworks. 
\item The proposed tool is made available as open-source software, which can play a crucial role in driving innovation in various DL-based CT applications.  
\item The proposed tool can be used for implementing analytical or iterative reconstruction algorithms, which facilitates an integration of DL-driven reconstructions with conventional optimization-based methods. 
\end{itemize}

% method
\begin{figure}[ht]
\begin{center}
\centerline{\includegraphics[width=0.8\columnwidth]{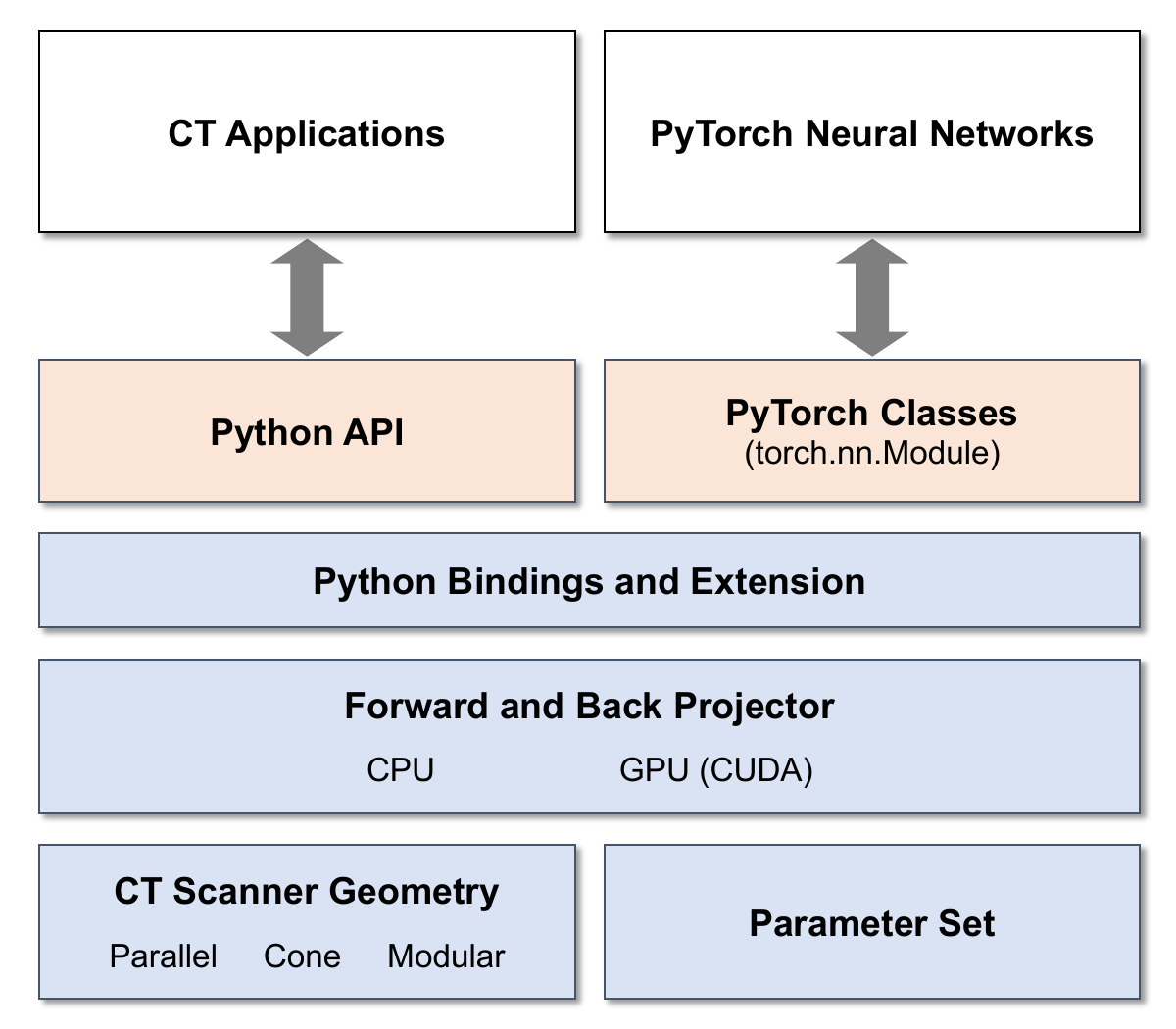}}
\caption{Overview of the proposed differentiable CT projector software architecture.}
\label{fig_overview}
\vspace{-0.5em}
\end{center}
\end{figure}

\section{Method}

The proposed differentiable CT projectors are based on the core C++/CUDA module, Python binding and PyTorch interfaces, as shown in Figure~\ref{fig_overview}. This section describes the projection models with the individual modules. 

\subsection{CT Projection Models and Geometries}

The XCT forward model is given by the X-ray Transform which is a collection of line integrals through the unknown 3D volume \cite{Nat86}. The XCT model can be formulated as $\ybm = \Abm\xbm$ where $\xbm \in \R^n$ is the volume of x-ray linear attenuation coefficients and $\ybm$ are the measured projection data. The forward model is denoted by $\Abm \in \R^{m \times n}$ where the coefficients model the measurement process with the geometry of the CT system. XCT reconstruction is an inverse problem to recover the volume $\xbm$. Accurate reconstruction requires knowledge of the 3D position of every x-ray source and x-ray detector pixel pair which describes the geometry of a single measurement.  Thus XCT forward models must be developed for the specific geometry of the imaging system.  We will now discuss the CT geometries that our software package supports and provide some details on the algorithms we implemented.

We chose to implement three different 3D XCT geometry types in this software package: parallel-beam, axial cone-beam (planar or curved detector), and a flexible cone-beam geometry where uses may place x-ray sources and detectors at any position and orientation in space.  These geometry types cover most XCT applications and future releases will include fan-beam and helical cone-beam geometries.

In practice, the reconstruction volume is composed of a 3D collection of voxels.  Several mathematical models have been developed that model the XCT line integrals through a voxelized volume.  The most popular and effective models are the Siddon method \cite{Siddon1985Fast}, Joseph method \cite{Joseph1982}, Distance Driven (DD) method \cite{Deman2004DD}, and the SF method \cite{Long2010SF}.  The second two of these methods, DD and SF, model the finite width of the detector pixels and volume voxels, while the first two of these methods do not.  Thus, although they are more computationally expensive to compute, the DD and SF methods are more accurate and other methods have been shown to produce artifacts in some cases \cite{Deman2004DD}.  We chose to implement the Siddon and SF projector methods in our software package.

Any optimization-based method that relies on the calculation of gradients requires one to calculate the adjoint (or in the discrete case, the transpose) of the X-ray Transform \cite{Barrett_ImageSci_2013}.  The adjoint of the X-ray Transform is commonly referred to as \textit{backprojection}.  With some caveats \cite{Zeng2000}, one should employ methods for backprojection that are the exact transpose of the forward projection, referred to as \textit{matched projectors}.  Although there are some XCT reconstruction packages that utilize match projectors \cite{Champley2022LTT}, most reconstruction packages \cite{ASTRA} violate this requirement because exact transposes are typically not as computationally efficient as other methods and if one stops the iterative reconstruction process early enough, artifacts will not appear.  Since our goal here is to implement methods that are stable after over a thousand or more iterations, we chose to implement methods where the exact transpose is used. As an example of why the adjoint is needed, consider the following least squares 
$ \widehat{x}_{LS} := \underset{x}{\operatorname{argmin}} \frac{1}{2}\|Ax - y\|^2. $
%\begin{eqnarray*}
%\widehat{x}_{LS} &:=& \underset{x}{\operatorname{argmin}} \frac{1}{2}\|Ax - y\|^2.
%\end{eqnarray*}
Then the gradient of this cost function is given by $ A^T(Ax - y). $

All of our projectors are quantitatively accurate.  The detector pixels and reconstruction voxels are specified in mm and the reconstruction volume units are in mm$^{-1}$.

Our parallel- and cone-beam implementations allow for a flexible specification of the geometry, including arbitrary 3D detector shifts and non-equispaced projection angles.  If these still do not offer a flexible enough geometry, the user can specify the 3D location of every source-detector pair, as well as an arbitrary orientation of the detector.

Lastly, we also implemented forward/ back projector pairs for objects with cylindrical symmetry \cite{ChampleyAbel}.  A special case of this is the Abel Transform which applies to parallel-beam geometries.

\subsection{Core CUDA Implementation}

Forward and back projection implementations are available for the three CT geometries described in the previous subsection using both the Siddon and SF projector models.  These algorithms are implemented for the CPU using C/C++ and for NVIDIA GPUs using CUDA.  Parallelization is done over the samples in the output space (CT projections for forward projection and reconstruction voxels for backprojection).  Our CUDA implementation utilizes 3D threads and 3D texture memory is used for the input data (reconstruction volume for forward projection and CT projections for backprojection).

CT projection data and the reconstruction volumes are both stored as contiguous 32-bit floating point arrays.  Users simply provide the software with pointers to the relevant arrays.  If the data are already on the GPU, then the appropriate method is implemented on this data, but if the data is not on the GPU (and the user want the calculation done on the GPU), the software copies the data to the GPU, performs the calculation, and then copies the result back to the CPU.

\subsection{PyTorch Interfaces and CT Parameters}

The forward and back projection in the core C++/CUDA module are provided as Python functions using the Python binding and the custom C++/CUDA extensions \cite{pytorch2019}. The main class is ``Projector'' which is derived from \textbf{torch.nn.Module} to enable automatic differentiation with PyTorch and the Python implementation. 

The forward and back projections in the proposed library require a set of parameters specifying the CT scanner system. These CT parameters include scanner geometry type, number of projections, angular range or list of projection angles, detector pixel width and heights, source-to-detector distance, source-to-object distance, horizontal/vertical detector shift, number of voxels in each dimension, voxel sizes, and volume center position. They can be specified using set functions or a configuration file.

% sample usage

\begin{figure}[t]
\begin{center}
%\centerline{\includegraphics[width=0.65\textwidth]{figures/fig_leap_app.pdf}}
\centerline{\includegraphics[width=0.99\columnwidth]{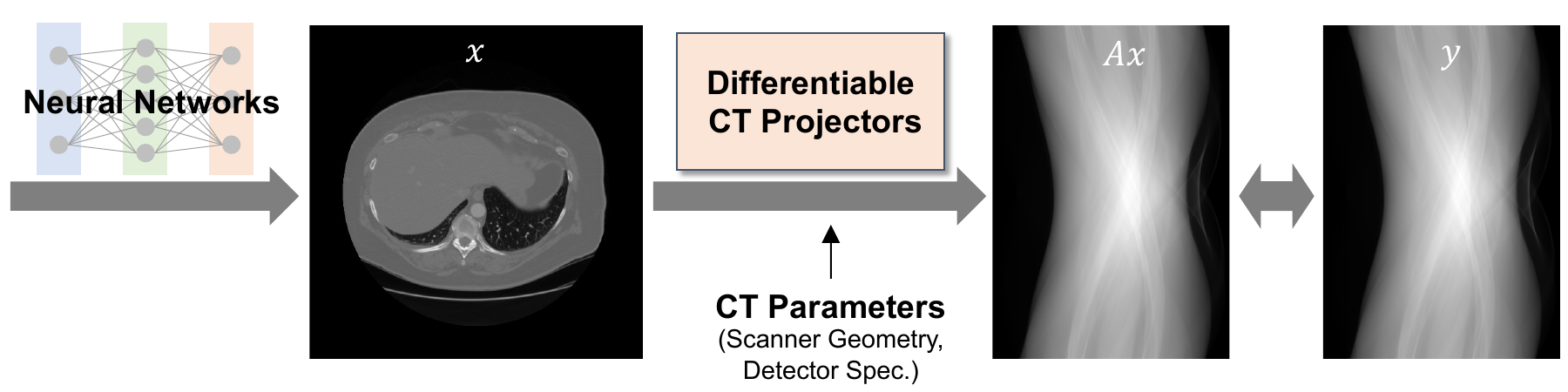}}
\caption{Model training and inference with our CT Projectors}
\label{fig_leap_app}
\vspace{-0.5em}
\end{center}
\end{figure}

\section{Sample Usage}

\paragraph{Integration with Model Training and Inference}

A few recent DL methods addressing ill-posed CT reconstructions use forward projection models for data consistency \cite{zhou2019LACT, liu2022DOLCE, lahiri2023}. These methods utilize a forward model as a loss function in the training or inference pipelines. While the detailed formulations differ from each other, the basic idea is to minimize the difference between the forward-projected data and original projection data, as illustrated in Figure~\ref{fig_leap_app}. The loss function can be formulated as: $ \argmin_{\xbm\in\R^n} |\Abm\xbm - \ybm\|_2^2 $. 

Our proposed software library can be utilized as a differentiable forward model in existing DL frameworks. The provided PyTorch class enables seamless integration into existing \textbf{torch.nn.Module} classes. The Python code snippet below illustrates how to utilize our projection library during training or inference time:

\begin{lstlisting}[language=Python, caption=Example of our projector with a PyTorch model]
import torch
from leaptorch import Projector

proj = Projector(use_gpu=True, gpu_device="cuda:0", batch_size=1)
proj.load_param("parallel_beam.cfg")

for ind in range(0, n_iters):
  X, Y = data # retrieve data
  X_pred = model(X) # forward pass
  Y_pred = proj(X_pred) # forward project
  loss = loss_func(Y, Y_pred)
\end{lstlisting}
\vspace{-0.5em}

\paragraph{End-to-End Reconstruction Pipeline}

While the proposed software library is primarily designed for forward modeling in neural network models, it can also be utilized for additional reconstruction processes or standalone reconstruction applications. The majority of the neural network inference-based methods in CT applications aims to improve ill-posed input images from conventional reconstruction algorithms (\eg, Filtered Backprojection (FBP)) by removing artifacts and noise \cite{han2018unet, zhou2019LACT, liu2022DOLCE}. In general, these initial reconstruction processes are performed using separate CT applications. Yet, the initial reconstruction process using our library has several advantages. First, it enables an integrated training pipeline with efficient memory usage. Second, this method enables to augment diverse, ill-posed input images given the training projection data. One can easily randomize  the angular range (limited-angle CT) or the number of views (few-view CT) to generate diverse ill-posed input images. 

Moreover, the predicted images from the inference models can be further improved by additional reconstruction algorithms \cite{kim2019FVCT}. Our tool can also facilitate the seamless implementation of an end-to-end pipeline for the sinogram completion \cite{anirudh2018LACT}.

% experiments

\section{Experiments}

%\paragraph{Validation with Existing Methods}
%The software library provides a PyTorch-based filtered back projection (FBP) reconstruction module (included in the demos). We validated the reconstructed images with the ones by another publicly available reconstruction software tool (LTT) \cite{Champley2022LTT}. 

\paragraph{Data Consistency with Inference Models}

\begin{figure}[t!]
\begin{center}
\centerline{\includegraphics[width=0.99\columnwidth]{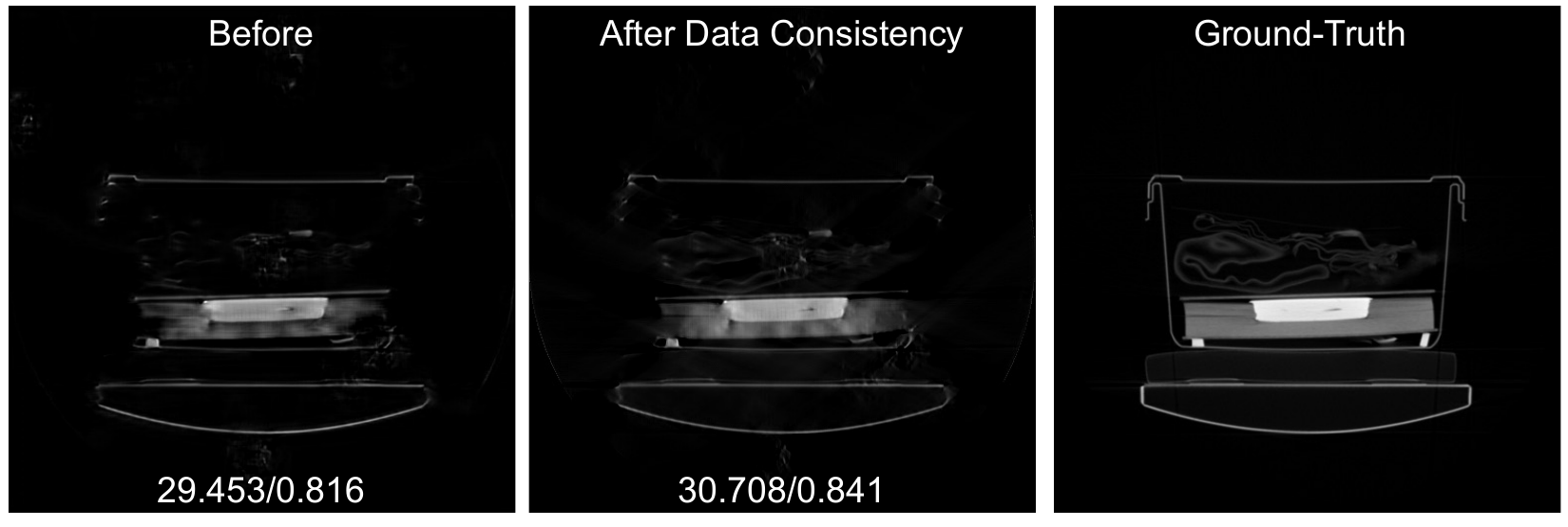}}
\caption{One of the predicted images from the inference model and its final image after the data consistency-based refinement with our projector. The PSNR/SSIM values are indicated at the bottom.}
\label{fig_result}
%\vspace{-0.5em}
\end{center}
\end{figure}

We performed limited-angle CT experiments to demonstrate the benefit of the data consistency with the proposed differentiable projectors. We used a publicly available airport luggage dataset for automatic threat recognition \cite{TO42014} that we split into 165 bags for training and the remaining 25 bags for test. The image dimension is $512^2$ and the number of projections is 720 (parallel beam). To demonstrate the limited-angle CT, we randomly masked $120^\circ$ out of $180^\circ$ ($60^\circ$ available). We implemented a neural network model combining CT-Net \cite{anirudh2018LACT} and U-Net \cite{han2018unet}. Once the model was trained, we performed the sinogram completion and the iterative data consistency step using our projectors during the inference time. We compared the predicted images from the inference model and the final images after the sinogram completion and iterative refinement, with the ground truth images. The signal-to-noise-ratio score (PSNR) and structural similarity (SSIM) were used as the image quality metric. Figure~\ref{fig_result} shows one of the predicted images directly from the model and its final image after the refinement. The refinement step with our projector led to an improvement in the averaged PSNR (dB) and SSIM from 35.486 and 0.905 to 36.350 and 0.911, respectively.

\paragraph{Performance and Memory Usage}

We report the forward projection time of our library and another publicly available reconstruction software tool (LTT) \cite{Champley2022LTT} on NVIDIA Tesla P100 with 16GB, as shown in Table~\ref{table_performance}. The angular ranges for parallel and cone beams are 180 and 360, respectively. Note that we did not list the memory usage for LTT because it is a user-specified parameter. The most GPU memory that LTT would use is enough to hold one copy of the projection data and volume data stored as 32-bit floats. This is also how much GPU memory LEAP requires as well.  This small memory footprint requirement enables easy integration with existing neural network models on GPUs.

%LTT stores all its data on the CPU and transfers memory on/off the GPUs as necessary. The user may specify how much GPU memory LTT is permitted to use. Then LTT transfers the data in small enough chunks on/off the GPU to perform a particular operation to stay within this user-specified value. 

%\iffalse
\begin{table}[t!]
  	\centering
  	\resizebox{\columnwidth}{!}
  	{\begin{tabular}{lllll}
	\toprule
 	\multicolumn{1}{l}{Geometry} & \multicolumn{2}{c}{Parallel} & \multicolumn{2}{c}{Cone}\\
Dimension & $512^3$/180 & $1024^3$/720 & $512^3$/180 & $1024^3$/720\\ 	
 	\hline\hline
Ours &  0.5/1.8 (1.5) &  11.5/15.4 (8) &  1.4/2.8 (1.5) &  37.1/39.2 (11.1) \\
LTT  &  4.2 (-) &  17.4 (-) &  4.5 (-)  & 38.9 (-) \\
\hline
\end{tabular}}
\caption{
	Performance comparison (sec) between ours and LTT. ($\cdot$) indicates the memory usage (GB). In our method, we report the times without and with the CPU-GPU data transfer. The dimension refers to the image dimension and the number of projections. 
 }
\label{table_performance}
\end{table}
%\fi

% conclusion

\section{Conclusion}

We presented an open-source software library providing differentiable XCT projectors. This library allows us to integrate forward modeling into DL frameworks as an end-to-end pipeline. This library enables fast forward and back projections with small memory footprint requirements.

\section*{Acknowledgments}
\small This work was performed under the auspices of the U.S. Department of Energy by Lawrence Livermore National Laboratory under Contract DE-AC52-07NA27344. LLNL-CONF-849839.

\bibliography{reference}
\bibliographystyle{icml2023}

%\newpage
%\appendix
%\onecolumn
%\section{You \emph{can} have an appendix here.}

\end{document}